\documentclass[12pt]{amsart}
\usepackage[english]{babel}
\usepackage{egothic}
\usepackage[T1]{fontenc}

\usepackage{amsmath}
\usepackage{amsthm}
\usepackage{amssymb}
\usepackage[all]{xy}
\usepackage{mathrsfs}
\usepackage{enumerate}
\usepackage{physics}
\usepackage[notcite, final, notref]{showkeys}
\usepackage{dsfont}
\usepackage[dvips]{color}
\newtheorem{theorem}{Theorem}[section]

\newtheorem{proposition}[theorem]{Proposition}

\newtheorem{definition}[theorem]{Definition}

\theoremstyle{definition}
\newtheorem{remark}[theorem]{Remark}

\newtheorem{example}[theorem]{Example}

\newcommand{\BC}{{\mathbb B}{\mathbb C}}
\newcommand{\D}{{\mathbb D}}

\renewcommand{\i}{{\bf i}}
\renewcommand{\j}{{\bf j}}
\renewcommand{\k}{{\bf k}}
\newcommand{\C}{{\mathbb C}}
\newcommand{\R}{{\mathbb R}}
\newcommand{\e}{{\bf e}}

\renewcommand{\Re}{\mathrm{Re}}

\usepackage{xcolor}

\title[Two BLMS One MLMS algorithms]{Two Bicomplex and One Multicomplex Least Mean Square algorithms}

\author[D. Alpay]{Daniel Alpay}
\address{(DA) Schmid College of Science and Technology \\
Chapman University\\
One University Drive
Orange, California 92866\\
USA}
\email{alpay@chapman.edu}

\author[K. Diki]{Kamal Diki}
\address{(KD) Schmid College of Science and Technology \\
Chapman University\\
One University Drive
Orange, California 92866\\
USA}
\email{diki@chapman.edu}

\author[M. Vajiac]{Mihaela Vajiac}
\address{(MV) Schmid College of Science and Technology \\
Chapman University\\
One University Drive
Orange, California 92866\\
USA}
\email{mbvajiac@chapman.edu}


\thanks{D. Alpay, K. Diki and M. Vajiac thank Champan University for Faculty Opportunity Fund which helped fund this research.\ D. Alpay also thanks the Foster G. and Mary McGaw Professorship in Mathematical Sciences, which supported his research}

\begin{document}
\maketitle
\begin{abstract}
  We study and introduce new gradient operators in the complex and bicomplex settings, inspired from the well-known Least Mean Square (LMS) algorithm invented in 1960 by Widrow and Hoff for Adaptive Linear Neuron (ADALINE).
  These gradient operators will be used to formulate new learning rules for the Bicomplex Least Mean Square (BLMS) algorithms and we will also formulate these learning rules will for the case of multicomplex LMS algorithms (MLMS). This approach extends both the classical real and complex LMS algorithms.
\end{abstract}

\section{Introduction}
\setcounter{equation}{0}

We study and introduce new gradient operators in the complex and bicomplex settings, inspired from the well-known Least Mean Square (LMS) algorithm invented in 1960 by Widrow and Hoff for Adaptive Linear Neuron (ADALINE).
In the past few decades, hypercomplex neural networks have started to become more popular, different machine learning techniques have been developed in the framework of hypercomplex numbers such as bicomplex numbers, quaternions and Clifford algebras, and it has been shown that such algorithms improve efficiency, see for example \cite{BS2008, JMTQG,  Kobayashi1, Kobayashi2, TMQLMS, PML2020, Valle2021}. It is worth mentioning that in other works, such as~\cite{Vieira1} the bicomplex neural networks were considered and a hypergeometric activation function was used, while a quaternionic counterpart of convolutional neural networks with Bessel activation functions were also studied in \cite{Vieira2}. Moreover, the authors of~\cite{Took} introduced a quaternionic least mean square algorithm (QLMS) for adaptive filtering of three and four dimensional processes and studied some properties of these QLMS algorithms.

In this research paper, inspired by the work of Brandwood, Widrow et al.~\cite{Brandwood, WMcB}, we propose to derive two bicomplex LMS algorithms extending both the classical real and complex LMS algorithms of Widrow, and extend on of these algorithm to a multicomplex LMS as well.  We would like to mention that, in their work~\cite{bcplx_perceptron}, the authors have also extended the complex perceptron algorithm to the bicomplex case, in an effort to extend the the framework of the perceptron theory to the bicomplex case, where the XOR controversy may be resolved.

We have not run simulations applications to demonstrate the advantages of the algorithms designed in this manuscript, however, in~\cite{Kobayashi1, Kobayashi2} a comparison between the speed and efficiency of such algorithms has been shown, for example Figures 3,4, and 5 in~\cite{Kobayashi1} and Table 1 in~\cite{Kobayashi2}. We posit that the same types of improved algorithmic speed will be prevalent in our case as well, as there is strong evidence in the literature that using hypercomplex analysis increase efficiencies, just as generalizing from the real case to the complex one does.

The paper is structured as follows: in Section~\ref{bcplx_grad}, following the same type of reasoning as in the complex case above, we use notions of bicomplex analysis to establish an extension of the Wirtinger (or CR) Calculus to this case. In Section~\ref{BLMS_alg}, we fulfill the purpose of this paper, i.e. we establish two types of bicomplex LMS algorithms, each based on a different type of conjugation and its respective gradient operator.Then, in Section~\ref{multicomplex-LMS} we use the bicomplex LMS algorithms considered in Section~\ref{BLMS_alg} to present the multicomplex counterpart of the BLMS algorithms.
The more involved theoretical aspects and details of Section~\ref{bcplx_grad} and~\ref{BLMS_alg} are placed at the end of the paper, starting with a reiteration in Appendix A of Brandwood's work~\cite{Brandwood}, who used complex CR-Calculus to  establish the complex LMS algorithm. In Appendix B we place the more theoretical proofs of the BCR (or bicomplex Wirtinger) calculus needed for the creation of the BLMS algorithms, while in Appendix C we place the proofs of the two BLMS algorithms we found.

For the interested reader, complementing this paper, the authors also wrote an extension of the BCR-Calulus to the bicomplex Hermite polynomials and bicomplex polyanalytic functions, which can be found in~\cite{ADV-BCR,ADV-BCR2}. 

\section{Bicomplex gradient operators}
\label{bcplx_grad}
The algebra of bicomplex numbers was first introduced by Segre in~\cite{Segre}. During the past decades, many mathematical works analyzed either the properties of bicomplex numbers, or the properties of holomorphic functions defined on bicomplex numbers, and, without pretense of completeness, we direct the attention of the reader first to the to book of Price,~\cite{Price}, where a full foundation of the theory of multicomplex numbers was given, then to some of the works describing some analytic properties of functions in the field~\cite{alss, CSVV, DSVV, mltcplx}. Applications of bicomplex (and other hypercomplex) numbers can be also found in the works of Alfsmann, Sangwine, Gl\"{o}cker, and Ell~\cite{AG, AGES}.\\

We now introduce, in the same fashion as~\cite{CSVV,bcbook,Price}, the key definitions and results for the case of holomorphic functions of complex variables. 
More details can be found in Appendix A.
The algebra of bicomplex numbers is generated by two commuting imaginary units $\i$
and $\j$ and we will denote the bicomplex space by $\BC$.  The product of the two commuting units  $\i$ and $\j$ is denoted by $ \k := \i\j$ and we note that $\k$ is a hyperbolic unit, i.e. it is a unit which squares to $1$.  Because of these various units in $\BC$, there are several
different conjugations that can be defined naturally. We will make use of these appropriate conjugations in this paper, and we refer the reader to~\cite{bcbook,mltcplx} for more information on bicomplex and multicomplex analysis. 
A bicomplex number can initially be written as  $Z=x_1+\i x_2+\j x_3 +\k x_4=z_1+\j z_2$, where $z_1=x_1+\i x_2$ and $z_2= x_3 +\j x_4$.
In the bicomplex algebra addition is component-wise and multiplication follows the algebraic rules $\i \j=\j\i=\k$.

\medskip
\subsection{Properties of the bicomplex algebra}

The bicomplex space, $\BC$, is not a division algebra, and it has two distinguished zero
divisors, $\e_1$ and $\e_2$, which are idempotent, linearly independent
over the reals, and mutually annihilating with respect to the
bicomplex multiplication:
\begin{align*}
  \e_1&:=\frac{1+\k}{2}\,,\qquad \e_2:=\frac{1-\k}{2}\,,\\
  \e_1 \cdot \e_2 &= 0,\qquad
  \e_1^2=\e_1 , \qquad \e_2^2 =\e_2\,,\\
  \e_1 +\e_2 &=1, \qquad \e_1 -\e_2 = \k\,.
\end{align*}
Just like $\{1,\mathbf{j} \},$ they form a basis of the complex algebra
$\BC$, which is called the {\em idempotent basis}. If we define the
following complex variables in $\C(\i)$:
\begin{align*}
  \beta_1 := z_1-\i z_2,\qquad \beta_2 := z_1+\i z_2\,,
\end{align*}
the $\C(\i)$--{\em idempotent representation} for $Z=z_1+\j z_2$ is
given by
\begin{align*}
  Z &= \beta_1\e_1+\beta_2\e_2\,.
\end{align*}

The $\C(\i)$--idempotent is the only representation for which
multiplication is component-wise, as shown in the next lemma.

\begin{remark}
  \label{prop:idempotent}
  The addition and multiplication of bicomplex numbers can be realized
  component-wise in the idempotent representation above. Specifically,
  if $Z= a_1\,\e_2 + a_2\,\e_2$ and $W= b_1\,\e_1 + b_2\,\e_2 $ are two
  bicomplex numbers, where $a_1,a_2,b_1,b_2\in\C(\i)$, then
  \begin{eqnarray*}
    Z+W &=& (a_1+b_1)\,\e_1  + (a_2+b_2)\,\e_2   ,  \\
    Z\cdot W &=& (a_1b_1)\,\e_1  + (a_2b_2)\,\e_2   ,  \\
    Z^n &=& a_1^n \,\e_1  + a_2^n \,\e_2  .
  \end{eqnarray*}
  Moreover, the inverse of an invertible bicomplex number
  $Z=a_1\e_1 + a_2\e_2 $ (in this case $a_1 \cdot a_2 \neq 0$) is given
  by
  $$
  Z^{-1}= a_1^{-1}\e_1 + a_2^{-1}\,\e_2 ,
  $$
  where $a_1^{-1}$ and $a_2^{-1}$ are the complex multiplicative
  inverses of $a_1$ and $a_2$, respectively.
\end{remark}

\begin{remark}
These split the bicomplex space in $\BC=\mathbb C \mathbf{e}_1\bigoplus \mathbb C \mathbf{e}_2$, as:
\begin{equation}
  Z=z_1+\j z_2=(z_1-\i z_2)\mathbf{e}_1+(z_1+\i z_2)\mathbf{e}_2=\lambda_1\e_1+\lambda_2\e_2.
\end{equation}
\end{remark}

Simple algebra yields:
\begin{equation}
\begin{split}
  z_1&=\frac{\lambda_1+\lambda_2}{2}\\
    z_2&=\frac{\i(\lambda_1-\lambda_2)}{2}.
  \end{split}
  \end{equation}

Because of these various units in $\BC$, there are several
different conjugations that can be defined naturally and we will now define the conjugates in the bicomplex setting, as in~\cite{CSVV,bcbook}

\begin{definition} For any $Z\in \BC$ we have the following three conjugates:
  \begin{eqnarray}
  \label{conj}
    \overline{Z}=\overline{z_1}+\j\overline{z_2},\\
     Z^{\dagger}=z_1-\j z_2,\\
      Z^*=\overline{Z^{\dagger}}=\overline{z_1}-\j\overline{z_2}.
  \end{eqnarray}
\end{definition}

\begin{remark}
\label{id-conj}
Moreover, following Definition~\ref{conj}, if we write $Z=\lambda_1 \mathbf{e}_1+\lambda_2\mathbf{e}_2$ in the idempotent representation,we have 
\begin{align*}
  Z &= \lambda_1 \e_1 + \lambda_2 \e_2  \\
 Z^\ast  &=  \overline{\lambda_1} \e_1 + \overline{\lambda_2} \e_2 \\
   Z ^\dagger&= \lambda_2 \e_1 + \lambda_1 \e_2  \\
   \overline{Z} &=  \overline{\lambda_2} \e_1 + \overline{\lambda_1} \e_2\,,
\end{align*}
\end{remark}
We refer the reader to~\cite{bcbook} for more details.
\bigskip


The Euclidean norm $\|Z\|$ on $\BC$, when it is seen as
$\C^2(\i), \C^2(\j)$ or $\R^4$ is:
\begin{align*}
  \|Z\| = \sqrt{ | z_1 | ^2 + | z_2 |^2 \, } = \sqrt{ \Re\left( | Z |_\k^2 \right) \, } = \sqrt{
  \, x_1^2 + y_1^2 + x_2^2 + y_2^2 \, }.
\end{align*}
As studied in detail in~\cite{bcbook}, in idempotent
coordinates $Z=\lambda_1\e_1+\lambda_2\e_2$, the Euclidean norm becomes:
\begin{align}
  \label{Euclidean_idempotent}
  \|Z\| = \frac{1}{\sqrt2}\sqrt{|\lambda_1|^2 + |\lambda_2|^2}.
\end{align}

It is easy to prove that
\begin{align}
  \|Z \cdot W\|  \leq  \sqrt{2} \left(\|Z\| \cdot  \|W\| \right).
\end{align}
\begin{definition}

One can define a
{\em hyperbolic-valued} norm for $Z=z_1+\j z_2 = \lambda_1 \e_1+\lambda_2\e_2$
by:
\begin{align*}
  \| Z\|_{\D_+} := |\lambda_1|\e_1 + |\lambda_1|\e_2 \in\D^+.
\end{align*}
\end{definition}
It is shown in~\cite{alss} that this definition obeys the corresponding properties
of a norm, i.e. $ \| Z\|_{\D_+}=0$ if and only if $ Z=0$,
it is multiplicative, and it
respects the triangle inequality with respect to the order introduced
above. 
In the paper~\cite{bcplx_perceptron} we have used an extension of the previous norm to the space of $\BC$ vectors, i.e. elements of $\BC^n$, to prove a perceptron theorem in the bicomplex setting.

In this section we briefly define and revise from \cite{ADV-BCR} basic properties of bicomplex gradient operators in the case of several bicomplex variables which will be applied in the sequel.

\subsection{Bicomplex differential operators}

Let us first recall that for a given bicomplex number $Z=z_1+\mathbf{j}z_2$ we have (for more details see~\cite{bcbook, Price}):
$$z_1=\frac{Z+Z^\dagger}{2}, \quad z_2=\frac{\mathbf{j}}{2}(Z^\dagger-Z),$$
and 
$$\overline{z_1}=\frac{\overline{Z}+Z^*}{2}, \quad \overline{z_2}=\frac{\mathbf{j}}{2}(Z^*-\overline{Z}).$$

It follows that
$$x_1=\frac{Z+Z^\dagger+Z^*+\overline{Z}}{4}, \quad x_2=\frac{\mathbf{i}}{4}(Z^*+\overline{Z}-Z-Z^\dagger);$$
and $$x_3=\frac{\mathbf{j}}{4}(Z^*+Z^{\dagger}-\overline{Z}-Z), \quad x_4=\frac{\mathbf{k}}{4}(Z+Z^*-\overline{Z}-Z^\dagger).$$

Following~\cite{CSVV}, the bicomplex differential operators with respect to the various bicomplex conjugates  are:

$$\partial_{Z}:= \partial_{z_1}-\mathbf{j} \partial_{z_2}=\partial_{x_1}-\mathbf{i}\partial{x_2}-\mathbf{j}\partial_{x_3}+\mathbf{k}\partial{x_4}=\partial_{\lambda_1}\mathbf{e}_1+\partial_{\lambda_2}\mathbf{e}_2;$$
 $$\partial_{\overline{Z}}:= \partial_{\overline{z_1}}-\mathbf{j} \partial_{\overline{z_2}}=\partial_{x_1}+\mathbf{i}\partial{x_2}-\mathbf{j}\partial_{x_3}-\mathbf{k}\partial{x_4}= \partial_{\overline{\lambda_2}}\mathbf{e}_1+\partial_{\overline{\lambda_1}}\mathbf{e}_2;$$
  $$\partial_{Z^*}:= \partial_{\overline{z_1}}+\mathbf{j} \partial_{\overline{z_2}}=\partial_{x_1}+\mathbf{i}\partial{x_2}+\mathbf{j}\partial_{x_3}+\mathbf{k}\partial{x_4}=\partial_{\overline{\lambda_1}}\mathbf{e}_1+\partial_{\overline{\lambda_2}}\mathbf{e}_2;$$
  $$\partial_{Z^\dagger}:=\partial_{z_1}-\mathbf{j} \partial_{z_2}=\partial_{x_1}-\mathbf{i}\partial{x_2}+\mathbf{j}\partial_{x_3}-\mathbf{k}\partial{x_4}=\partial_{\lambda_2}\mathbf{e}_1+\partial_{\lambda_1}\mathbf{e}_2.$$ 


 \subsection{Properties of bicomplex gradient operators}
 \label{bcplx_grad_op}
 
In Appendix B we review the concept of BCR-analytic functions to functions of several bicomplex variables introduced in~\cite{ADV-BCR} and here we define the resulting gradient operators for such functions. 
We define the gradient operators corresponding to the $Z^*$ and $\overline{Z}$ conjugations, $\nabla_{\mathsf{Z}^*}$ and $\nabla_{\overline{\mathsf{Z}}}$ respectively, which we will use to prove our bicomplex LMS algorithms.

 \begin{definition}[Bicomplex gradient operator]
 Let $\mathsf{Z}=\mathbf{z_1}+\mathbf{j}\mathbf{z_2}=\Lambda_1 \mathbf{e}_1+\Lambda_2 \mathbf{e}_2 \in \mathbb{BC}^n$ where $\mathsf{Z}=(Z_1,\dots,Z_n)$, with complex vector components $\mathbf{z}_1=(z_{11},...,z_{1n}),$ $\mathbf{z}_2=(z_{21},...,z_{2n}) $ and $ \Lambda_1=(\lambda_{11},...,\lambda_{1n}), \quad \Lambda_2=(\lambda_{21},...,\lambda_{2n})$ belong to $\mathbb{C}^n$. Then, we define the various bicomplex gradient operators with respect to the variables $\mathsf{Z}, \overline{\mathsf{Z}}$ and $\mathsf{Z}^*$ by:  
 
\begin{itemize}
\item[i)] Bicomplex gradient operatror:
$$\nabla_{\mathsf{Z}}:= \nabla_{\mathbf{z_1}}-\mathbf{j} \nabla_{\mathbf{z_2}}=\nabla_{\Lambda_1}\mathbf{e}_1+\nabla_{\Lambda_2}\mathbf{e}_2;$$
\item[ii)] Bicomplex gradient-bar operator: 
 $$\nabla_{\overline{\mathsf{Z}}}:= \nabla_{\overline{\mathbf{z_1}}}-\mathbf{j} \nabla_{\overline{\mathbf{z_2}}}= \nabla_{\overline{\Lambda_2}}\mathbf{e}_1+\nabla_{\overline{\Lambda_1}}\mathbf{e}_2;$$
 \item[iii)] Bicomplex gradient-$*$ operator:
 $$\nabla_{\mathsf{Z}^*}:= \nabla_{\overline{\mathbf{z_1}}}+\mathbf{j} \nabla_{\overline{\mathbf{z_2}}}=\nabla_{\overline{\Lambda_1}}\mathbf{e}_1+\nabla_{\overline{\Lambda_2}}\mathbf{e}_2.$$

\end{itemize}

 \end{definition}
We start by enumerating various interesting properties for these differential operators, adding some proofs in Appendix B.

\begin{theorem}[Bicomplex-gradient Leibniz rules]\label{LeibBCn}
Let $f$ and $g$ be two bicomplex-valued $BC$-R analytic functions. Then, it holds that 
\begin{equation}\label{G1}
\nabla_{\mathsf{Z}}(fg)=f\nabla_{\mathsf{Z}}(g)+ \nabla_{\mathsf{Z}}(f) g,
\end{equation}

\begin{equation}\label{G2}
\nabla_{\overline{\mathsf{Z}}}(fg)=f\nabla_{\overline{\mathsf{Z}}}(g)+ \nabla_{\overline{\mathsf{Z}}}(f) g,
\end{equation}
and 
\begin{equation}\label{G3}
\nabla_{\mathsf{Z}^*}(fg)=f\nabla_{\mathsf{Z}^*}(g)+ \nabla_{\mathsf{Z}^*}(f) g.
\end{equation}

\end{theorem}

The proofs of the following basic properties of bicomplex gradient operators are left to the reader:
\begin{proposition}
Let $\mathsf{Z}, \mathsf{a}\in  \mathbb{BC}^n $ and $\mathsf{R}$ be in $\mathbb{BC}^{n\times n}$ then the following properties hold true
\begin{enumerate}
\item $\nabla_{\overline{\mathsf{Z}}}(\mathsf{\overline{a}}^T\mathsf{Z})=0,$
\item $\nabla_{\overline{\mathsf{Z}}}(\mathsf{\overline{Z}}^T \mathsf{a})=\mathsf{a},$
\item $\nabla_{\overline{\mathsf{Z}}}(\mathsf{\overline{Z}}^T \mathsf{R} \mathsf{Z})=RZ,$
\item $\nabla_{\mathsf{Z}^*}(\mathsf{\overline{a}}^T\mathsf{Z})=0,$
\item $\nabla_{\mathsf{Z}^*}(\mathsf{(Z^*)}^T \mathsf{a})=\mathsf{a},$
\item $\nabla_{\mathsf{Z}^*}(\mathsf{(Z)^*}^T \mathsf{R} \mathsf{Z})=RZ.$
\end{enumerate}
\end{proposition}

\begin{remark}
Our paper does not enter in the theory of bicomplex holomorphic functions, we only need the gradient operators for the results presented here. The reader interested in the bicomplex and multicomplex analytic theories is invited to read~\cite{alss,bcbook,mltcplx}, for example.
\end{remark}

\section{Bicomplex LMS algorithms}
\label{BLMS_alg}

Let $\mathsf{Y}_\ell$ be the actual bicomplex output and $\mathsf D_\ell$ denote the desired output at time $\ell$. We note that the actual output is defined by 
$$\mathsf{Y}_\ell=\mathsf{X}_\ell^T \mathsf{W}_\ell=\mathsf{W}_\ell^T \mathsf{X}_\ell=\displaystyle \sum_{k=1}^n\mathsf{X}_{\ell, k} \mathsf{W}_{\ell, k},$$
where the bicomplex vectors $\mathsf{X}_\ell$ and $\mathsf{W}_\ell$ in $\mathbb{BC}^n$ are given repectively by $\mathsf{X}_\ell=(\mathsf{X}_{\ell,1},...,\mathsf{X}_{\ell,n})$ and $\mathsf{W}_\ell=(\mathsf{W}_{\ell,1},...,\mathsf{W}_{\ell,n})$. Finally, we introduce the bicomplex-valued error signal at time $\ell$ which is defined to be 
$$\mathsf{E}_\ell:=\mathsf{D}_\ell-\mathsf{Y}_\ell.$$

In the next subsections we will introduce and prove two bicomplex extensions of the well-known LMS algorithm and some related results.
\subsection{First Bicomplex Least Mean Square (BLMS) algorithm}
We propose here the first bicomplex LMS algorithm which can be introduced thanks to the use of the bicomplex $*$ conjugate instead of the complex conjugate as in the classical paper of Widrow et.al. This allows to obtain a first extension of the complex LMS algorithm to the bicomplex setting.

\begin{definition}[First BLMS learning rule]
We define the first bicomplex LMS algorithm by the following learning rule
\begin{equation}\label{bicompW1}
\mathsf{W}_{\ell+1}=\mathsf{W_\ell}-\mu\nabla_{\mathsf{W}_\ell^*}(\mathsf{E}_\ell \mathsf{E}_\ell^*),
\end{equation}
where $\mathsf{E}_\ell$ is the bicomplex signal error, $*$ is the bicomplex conjugate and $\mu>0$ is a real constant.
\end{definition}
In the next result we prove that the first bicomplex LMS algorithm can be derived by applying the bicomplex gradient operator $\nabla_{\mathsf{Z}^*}$. We only state the theorem and prove in in Appendix C of this work.

\begin{theorem}[First BLMS algorithm]
\label{LMSR1}
The learning rule at time $\ell$ of the first BLMS algorithm has the following explicit expression:
 \begin{equation}
\mathsf{W}_{\ell+1}=\mathsf{W_\ell}+2\mu \mathsf{E}_\ell\mathsf{X}_\ell^*.
\end{equation}
\end{theorem}

In the next results we express the learning rule for the first BLMS algorithm in terms of two classical complex LMS algorithms using two bicomplex decompositions. We also delegate the proof of this Theorem to Appendix C.

\begin{theorem}\label{LMS1decomp}
Let us consider the decomposition of the bicomplex weights, error and input which are given respectively by $$\mathsf{W}_\ell=w_{\ell,1}\mathbf{e}_1+w_{\ell,2}\mathbf{e}_2, \mathsf{E}_\ell=e_{\ell,1}\mathbf{e}_1+e_{\ell,2}\mathbf{e}_2,\quad \text{ and }\mathsf{X}_\ell=x_{\ell,1}\mathbf{e}_1+x_{\ell,2}\mathbf{e}_2.$$ Then, at time $\ell$ the learning rule of the first bicomplex LMS algorithm can be expressed in terms of two complex LMS algorithms. 
\begin{align*}
\mathsf{W}_{\ell+1}&=(w_{\ell,1}+2\mu e_{\ell,1}\overline{x_{\ell,1}})\mathbf{e}_1+(w_{\ell,2}+2\mu e_{\ell,2}\overline{x_{\ell,2}})\mathbf{e}_2\\
&=w_{\ell+1,1}\mathbf{e}_1+w_{\ell+1,2}\mathbf{e}_2.
\end{align*}
\end{theorem}

\begin{proposition}
Let us consider the decomposition of the bicopmplex weights, error and input at time $\ell$ given by $$\mathsf{W}_\ell=W_{\ell,1}+W_{\ell,2}\mathbf{j}, \mathsf{E}_\ell=E_{\ell,1}+E_{\ell,2}\mathbf{j},\quad \text{ and }\mathsf{X}_\ell=X_{\ell,1}+X_{\ell,2}\mathbf{j}.$$ Then, the learning rule of the first bicomplex LMS algorithm can be expressed in terms of two complex LMS algorithms as follows: 
\begin{equation}
\mathsf{W}_{\ell+1}=\left(W_{\ell,1}+2\mu (E_{\ell,1}\overline{X_{\ell,1}}+\mathsf{E_{\ell,2}}\overline{\mathsf{X}_{\ell,2}})\right)+\mathbf{j}\left(W_{\ell,2}+2\mu (E_{\ell,2}\overline{X_{\ell,1}}-\mathsf{E}_{\ell,1}\overline{\mathsf{X}_{\ell,2}})\right).
\end{equation}
\end{proposition}
\begin{proof}
We apply the first BLMS algorithm obtained in Theorem \ref{LMSR1} and the bicomplex decomposition to obtain

\begin{align*}
\mathsf{W}_{\ell+1}&=\mathsf{W_\ell}+2\mu\mathsf{E}_\ell\mathsf{X}_\ell^*\\
&=\left(W_{\ell,1}+W_{\ell,2}\mathbf{j}\right)+2\mu (E_{\ell,1}+E_{\ell,2}\mathbf{j})(\overline{X_{\ell,1}}-\overline{X_{\ell,2}}\mathbf{j})\\
&=W_{\ell,1}+W_{\ell,2}\mathbf{j}+2\mu (E_{\ell,1}\overline{\mathsf{X}_{\ell,1}}-E_{\ell,1}\overline{\mathsf{X}_{\ell,2}}\mathbf{j}+E_{\ell,2}\overline{X_{\ell,1}}\mathbf{j}+\mathsf{E}_{\ell,2}\overline{X_{\ell,2}}) \\
&=\left(W_{\ell,1}+2\mu (E_{\ell,1}\overline{X_{\ell,1}}+\mathsf{E_{\ell,2}}\overline{\mathsf{X}_{\ell,2}})\right)+\mathbf{j}\left(W_{\ell,2}+2\mu (E_{\ell,2}\overline{X_{\ell,1}}-\mathsf{E}_{\ell,1}\overline{\mathsf{X}_{\ell,2}})\right).\\
\end{align*}
This ends the proof.
\end{proof}

\subsection{Second Bicomplex Least Mean Square (BLMS) algorithm}
 
 This part is reproduced by analogy with the previous subsection by taking the $bar$-bicomplex conjugate in the learning rule rather than the $*$-conjugate. Some proofs are based on similar arguments used for the first BLMS, so we omit to give all the details. This allows to obtain a second extension of the complex LMS algorithm invented by Widrow et.al. in~\cite{WMcB} to the bicomplex setting.
\begin{definition}[Second BLMS learning rule]\label{LMSR2}
We define the second bicomplex LMS algorithm by the following learning rule
\begin{equation}\label{bicompW}
\mathsf{W}_{\ell+1}=\mathsf{W_\ell}-\mu \nabla_{\overline{\mathsf{W}_\ell}}(\mathsf{E}_\ell\overline{\mathsf{E}_\ell}),
\end{equation}
where $\mathsf{E}_\ell$ is the error, $-$ is the bicomplex $bar$-conjugate and $\mu>0$ is a real constant.
\end{definition}
In the next result we show that the second bicomplex LMS algorithm introduced in Definition \ref{LMSR2} can be obtained by applying the bicomplex gradient operator $\nabla_{\overline{\mathsf{Z}}}$. We delegate this proof to Appendix C as well.

\begin{theorem}[Second BLMS algorithm]
\label{SLMSalgo}
The learning rule at time $\ell$ of the second bicomplex LMS algorithm has the following explicit expression: \begin{equation}
\mathsf{W}_{\ell+1}=\mathsf{W_\ell}+2\mu \mathsf{E}_\ell\overline{\mathsf{X}_\ell}.
\end{equation}
\end{theorem}

In the next results we express the learning rule for the second bicomplex LMS algorithm in terms of two complex LMS algorithms based on the two bicomplex decompositions.

\begin{theorem}\label{LMS2decomp}
Let us consider the decomposition of the bicopmplex weights, error and input which are given respectively by $$\mathsf{W}_\ell=w_{\ell,1}\mathbf{e}_1+w_{\ell,2}\mathbf{e}_2, \quad \mathsf{E}_\ell=e_{\ell,1}\mathbf{e}_1+e_{\ell,2}\mathbf{e}_2,\quad \text{ and }\quad \mathsf{X}_\ell=x_{\ell,1}\mathbf{e}_1+x_{\ell,2}\mathbf{e}_2.$$ Then, at time $\ell$ the learning rule of the second BLMS algorithm can be expressed in terms of two complex LMS algorithms as follows: 

\begin{align*}
\mathsf{W}_{\ell+1}&=(w_{\ell,1}+2\mu e_{\ell,1}\overline{x_{\ell,2}})\mathbf{e}_1+(w_{\ell,2}+2\mu e_{\ell,2}\overline{x_{\ell,1}})\mathbf{e}_2\\
&=w_{\ell+1,1}\mathbf{e}_1+w_{\ell+1,2}\mathbf{e}_2.
\end{align*}
\end{theorem}

\begin{proposition}
Let us consider the decomposition of the bicopmplex weights, error and input at time $\ell$ given by $$\mathsf{W}_\ell=W_{\ell,1}+W_{\ell,2}\mathbf{j}, \quad \mathsf{E}_\ell=E_{\ell,1}+E_{\ell,2}\mathbf{j},\quad \text{ and }\quad \mathsf{X}_\ell=X_{\ell,1}+X_{\ell,2}\mathbf{j}.$$ Then, the learning rule of the second bicomplex LMS algorithm can be expressed in terms of two complex LMS algorithms as follows: 
\begin{equation}
\mathsf{W}_{\ell+1}=\left(W_{\ell,1}+2\mu(E_{\ell,1}\overline{X_{\ell,1}}-\mathsf{E_{\ell,2}}\overline{\mathsf{X}_{\ell,2}})\right)+\mathbf{j}\left(W_{\ell,2}+2\mu (E_{\ell,2}\overline{X_{\ell,1}}+\mathsf{E}_{\ell,1}\overline{\mathsf{X}_{\ell,2}})\right).
\end{equation}
\end{proposition}
\begin{proof}
The computations follow similar arguments as in the first BLMS algorithm using the bicomplex $bar$-conjugation instead of the $*$-conjugation.
\end{proof}


\section{Multicomplex LMS Algorithm}
\label{multicomplex-LMS}

In this section we extend one of the results of the previous one to the multicomplex case, where the use of the idempotent representation is essential. In this case, using the single variable theory of the multicomplex case to build an elaborate algorithm that encompasses so much more will be very advantageous in terms of computation speed as it reduces the number of iterations that such an algorithm needs.

In~\cite{mltcplx}, one of the co-authors introduces algebraic and analytic techniques on multicomplex spaces, which we will now use to introduce a multicomplex LMS algorithm in this context.
\subsection{Introduction to Multicomplex Spaces}

Let $\i_1, \i_2,\dots,\i_n$ be imaginary units,
i.e. $\i_\iota^2=-1$. Then we define recursively the space
$$
\BC_n:=\{Z_n=Z_{n-1,1}+\i_nZ_{n-1,2}\,\big|\,Z_{n-1,1},Z_{n-1,2}\in\BC_{n-1}\}
$$
with the usual addition and multiplication. Recursively, we get:
$$
Z_n=\sum_{|I|=n-k}\,\prod_{t=k+1}^{n} (\i_t)^{\alpha_t-1} Z_{k,I}
$$
where $Z_{k,I}\in\BC_k$, $I=(\alpha_{k+1},\dots,\alpha_n)$, and
$\alpha_j\in\{1,2\}$.

We define the conjugation as follows:
$$
\overline{Z_n}^{\i_l} = \left\{\begin{array}{cc}
    \displaystyle\sum_{|I|=n-k}\,\displaystyle\prod_{\stackrel{t=k+1}{t\neq
        l}}^{n}
    (\i_t)^{\alpha_t-1}(-\i_l)^{\alpha_l-1} Z_{k,I} & \text{ if } l>k\\
    \displaystyle\sum_{|I|=n-k}\,\displaystyle\prod_{\stackrel{t=k+1}{t\neq
        l}}^{n} (\i_t)^{\alpha_t-1} \overline{Z_{k,I}}^{\i_l} & \text{
      if } l<k
  \end{array}\right.
$$
and
$$
\overline{Z_n}^{\i_1\dots\i_l} =
\overline{\overline{\overline{Z_n^{\i_1}}}^{\i_2\ldots}}^{\i_l}
$$
It turns out that
$$
\overline{Z_n}^{\i_l} = \sum_{|I|=n-k}\,\sum_{i=k+1}^n \delta_{l,i}(-1)^{\alpha_l-1}c_k Z_{k,I}
+ \sum_{|I|=n-k}\,\sum_{i=1}^k \delta_{l,i} c_k \overline{Z_{k,I}}^{\i_l}.
$$

\subsection{Idempotent Representation}

The idempotents (different than $0$ and $1$) are then organized as
folows. Denote by
\begin{eqnarray*}
  \e_{kl} &:=& \frac{1+\i_k\i_l}{2}\\
  \bar\e_{kl} &:=& \frac{1-\i_k\i_l}{2}.
\end{eqnarray*}
Then consider the following sets:
\begin{eqnarray*}
  S_1 &:=& \{\e_{n-1,n}, \bar\e_{n-1,n}\},\\ 
  S_2 &:=& \{\e_{n-2,n-1}\cdot S_1, \bar\e_{n-2,n-1}\cdot S_1\},\\ 
  &\vdots&\\
  S_{n-1} &:=& \{\e_{12}\cdot S_{n-2}, \bar\e_{12}\cdot S_{n-2}\}.\\ 
\end{eqnarray*}
At each stage $k$, the set $S_k$ has $2^k$ idempotents.

Note the following peculiar identities:
\begin{eqnarray*}
  \i_k\cdot\e_{kl} &=& \frac{\i_k-\i_l}{2} = -\frac{\i_l-\i_k}{2} = -\i_l\cdot\e_{kl}\\
  \i_k\cdot\bar\e_{kl} &=& \frac{\i_k+\i_l}{2} = \i_l\cdot\bar\e_{kl}
\end{eqnarray*}

\begin{theorem}
  In each set $S_k$, the product of any two idempotents is zero.
\end{theorem}
\begin{theorem}
  Any $Z_n\in\BC_n$ can be written as:
  $$
  Z_n=\sum_{j=1}^{2^k} Z_{n-k,j}\e_j,
  $$
  where $Z_{n-k,j}\in\BC_{n-k}$ and $\e_j\in S_k$.
\end{theorem}

Note that $\BC_0=\R$, $\BC_1=\C$, and $\BC_2=\BC$ and, in particular, we note:
\begin{remark}
In particular, for $k=n-1$, any $Z_n\in\BC_n$ is written as
\begin{equation}
  \label{sophie}
  Z_n=\Lambda_1\e_1 + \Lambda_2\e_2 +\cdots + \Lambda_{2^{n-1}}\e_{2^{n-1}}\,,
\end{equation}
where $\zeta_j\in\C_{\i_1}$.
\end{remark}

For a function defined on an open set $D\subset\BC_n$ (which restricts to an open set in $\C$ in each complex component), one can define the notion of $MC$-R analytic functions in the same way as the $BC$-R analytic functions in Appendix B. The study of $MC$-R analytic functions is now a work in progress, however, for simplicity, we will consider them here to be functions $F:D\longrightarrow \BC_n$ $F=(f_1,f_2,\cdots, f_{2^{n-1}})$ such that $f_k$ are C-R analytic in all $\lambda_l$ variables, for $1\le l,k \le 2^{n-1}$. For such functions we can define the gradient operator as follows.

 \begin{definition}[Multicomplex gradient operator]
 Let $\mathsf{Z}=\Lambda_1\e_1 + \Lambda_2\e_2 +\cdots + \Lambda_{2^{n-1}}\e_{2^{n-1}}$ where the multicomplex vector $\mathsf{Z}=(Z_1,\dots,Z_m)$, has complex vector components in the idempotent decomposition $ \Lambda_k=(\lambda_{k1},...,\lambda_{km}), \quad 1\le k \le 2^{n-1}$ which belong to $\mathbb{C}^m$. Then, we define the multicomplex $*-$gradient operators with respect to the conjugation $\mathsf{Z}^*=\overline{\Lambda_1}\e_1 + \overline{\Lambda_2}\e_2 +\cdots + \overline{\Lambda_{2^{n-1}}}\e_{2^{n-1}}$ by:  
 
\begin{equation}
\label{grad-multicplx}\nabla_{\mathsf{Z}^*}:= \nabla_{\overline{\Lambda_1}}\mathbf{e}_1+\nabla_{\overline{\Lambda_2}}\mathbf{e}_2+\cdots +  \nabla_{\overline{\Lambda_{2^{n-1}}}}\mathbf{e}_{2^{n-1}} .
\end{equation}

 \end{definition}

\begin{theorem}[Multicomplex-gradient Leibniz rules]\label{Leib-MC_n}
Let $f$ and $g$ be two multicomplex-valued $MC$-R analytic functions. Then, it holds that 
\begin{equation}\label{MC-L}
\nabla_{\mathsf{Z}^*}(fg)=f\nabla_{\mathsf{Z}^*}(g)+ \nabla_{\mathsf{Z}^*}(f) g.
\end{equation}

\end{theorem}

Analogous properties of the Multicomplex gradient follows and we can write a multicomplex LMS algorithm. We have:
\begin{proposition}
Let $\mathsf{Z}, \mathsf{a}\in  \mathbb{BC}_n^m $ and $\mathsf{R}$ be in $\mathbb{BC}_n^{m \times m}$ then the following properties hold true
\begin{enumerate}
\item $\nabla_{\mathsf{Z}^*}(\mathsf{a^*}^T\mathsf{Z})=0,$
\item $\nabla_{\mathsf{Z}^*}(\mathsf{(Z^*)}^T \mathsf{a})=\mathsf{a},$
\item $\nabla_{\mathsf{Z}^*}(\mathsf{(Z)^*}^T \mathsf{R} \mathsf{Z})=RZ.$
\end{enumerate}
\end{proposition}

\begin{remark}
Our paper does not enter in the theory of bicomplex holomorphic functions, we only need the gradient operators for the results presented here. The reader interested in the bicomplex and multicomplex analytic theories is invited to read~\cite{alss,bcbook,mltcplx}, for example.
\end{remark}

\begin{definition}[First MLMS learning rule]
We define the first multicomplex LMS algorithm by the following learning rule
\begin{equation}\label{multicompW1}
\mathsf{W}_{\ell+1}=\mathsf{W_\ell}-\mu\nabla_{\mathsf{W}_\ell^*}(\mathsf{E}_\ell \mathsf{E}_\ell^*),
\end{equation}
where $\mathsf{E}_\ell$ is the multicomplex signal error, $*$ is the multicomplex conjugate and $\mu>0$ is a real constant.
\end{definition}
In the next result we prove that the first bicomplex LMS algorithm can be derived by applying the bicomplex gradient operator $\nabla_{\mathsf{Z}^*}$. 

\begin{theorem}[First MLMS algorithm]
\label{MLMS1}
The learning rule at time $\ell$ of the first MLMS algorithm has the following explicit expression:
 \begin{equation}
\mathsf{W}_{\ell+1}=\mathsf{W_\ell}+2\mu \mathsf{E}_\ell\mathsf{X}_\ell^*.
\end{equation}
\end{theorem}

The proof is similar to the bicomplex one found in Appendix C of this work.


In the future, we will use the arguments considered using the complex gradient operator and study if they will still work in the quaternionic and hyperbolic cases. In particular, we want to propose a new least mean square algorithms (LMS) in these new hyperbolic and quaternionic settings, as well as a proof of the perceptron theorem in these cases.  


\section*{Appendix A: The Complex LMS Algorithm}
\label{Complex CR and LMS}

\setcounter{equation}{0}
\subsection*{C-R Calculus in the complex case}
\label{Cplx CR}

In the papers~\cite{Brandwood, Kreutz} an application of CR calculus to complex neural networks is discussed and we elaborate on some of the finer points of the theory in this subsection.

We observe that if a function $f(z)$ is real valued, we obtain $$\overline{\left(\frac{\partial}{\partial z}f(z)\right)}=\frac{\partial}{\partial \overline{z}}f(z),$$
and  it is clear that if $f$ belongs to the kernel of $\displaystyle \frac{\partial}{\partial \overline{z}}$ it will be in the kernel of $\displaystyle \frac{\partial}{\partial z}$ and vice versa.

We now introduce the following class of functions which can be considered the basis to study the C-R calculus (see~\cite{Brandwood, Kreutz}).
\begin{definition}
We  say that a function $f:\mathbb{C}\longrightarrow \mathbb{C}$ is \textbf{C-R analytic (or C-R regular)} if there exists a complex analytic function of two complex variables $g:\mathbb{C}\times \mathbb{C}\longrightarrow \mathbb{C}, (z_1,z_2)\mapsto g(z_1,z_2)$ such that 
\begin{itemize}
\item[i)] $f(z)=g(z,\overline{z}), \quad \forall z\in \mathbb{C}$,
\\
\item[ii)] $\displaystyle \frac{\partial f}{ \partial z}=\left(\frac{\partial g}{\partial z_1} \right)_{z_1=z, \,z_2=\overline{z}}$ ,
\\
\item[iii)]  $\displaystyle \frac{\partial f}{ \partial \overline{z}}=\left(\frac{\partial g}{\partial z_2} \right)_{z_1=z, \, z_2=\overline{z}}.$
\end{itemize}
\end{definition}
\begin{example}
\begin{enumerate}
\item The function $\displaystyle f(z)=\frac{z+\overline{z}}{2}$ is a real valued C-R analytic function with $\displaystyle g(z_1,z_2)=\frac{z_1+z_2}{2}.$ 
\item The function $f(z)=z^2\overline{z}$ is a complex valued C-R analytic function with $g(z_1,z_2)=z_1^2z_2$.
\end{enumerate}

\end{example}
\begin{proposition}
Any polyanalytic function of any order $n$ is C-R analytic.
\end{proposition}
\begin{proof}
Let $f:\mathbb{C}\longrightarrow \mathbb{C}$ be a polyanalytic function of order $n=1,2,...$. We know by poly-decomposition that there exists unique analytic functions $f_0,....,f_{n-1}$ such that 
$$f(z)=\displaystyle \sum_{k=0}^{n-1}\overline{z}^kf_k(z), \quad \forall z\in \mathbb{C}.$$
Then, we consider the analytic function of two complex variables defined by $$g(z_1,z_2)=\displaystyle \sum_{k=0}^{n-1}z_2^kf_k(z_1), \quad \forall (z_1,z_2)\in \mathbb{C}^2.$$ 
It is clear that $g(z,\overline{z})=f(z)$ and then we can easily check that $f$ is C-R analytic.
\end{proof}
\begin{example}
We consider the function $f(z)=e^{|z|^2}$. It is clear that $f$ is C-R analytic with $$g(z,w):=\displaystyle \sum_{n=0}^\infty \frac{z^n w^n}{n!}=e^{zw},\quad \forall z,w\in \mathbb{C}.$$
However, $f$ is not polyanalytic of a finite order.
\end{example}
\begin{remark}
The following formulas hold in the complex case
$$\frac{\partial |z|^{2k}}{ \partial \overline{z}}=kz|z|^{2k-2}, \quad \frac{\partial |z|^{2k+1}}{ \partial \overline{z}}=\frac{2k+1}{2}z|z|^{2k-1}, \quad k\ge 0 .$$
\end{remark}

In classic complex analysis we see that any function $f:\mathbb{C} \longrightarrow \mathbb{C}$ can be seen as a function from $\mathbb{R}\times\mathbb{R}$ to  $\mathbb{C}$ as $f(z)=f(x+ iy)=f(x,y)$. Here we re-formulate results of Brandwood in~\cite{Brandwood}.

\begin{theorem} Let $f$ be as above. If there exists $g:\mathbb{C}\times \mathbb{C}\longrightarrow \mathbb{C}$ a function which is analytic with respect to each of the two variables independently such that $$f(x,y)=g(z,\overline{z}), \quad z=x+iy.$$ Then, the differential operators follow the expected rules:  
\begin{eqnarray*} 
\frac{\partial g}{\partial z}&=\frac{1}{2}\left(\frac{\partial}{\partial x}-i\frac{\partial}{\partial y}\right)f(x,y),\\
 \frac{\partial g}{\partial \overline{z}}&=\frac{1}{2}\left(\frac{\partial}{\partial x}+i\frac{\partial}{\partial y}\right)f(x,y),
 \end{eqnarray*}

where $\displaystyle \frac{\partial g}{\partial z}:=\left(\frac{\partial g}{\partial z_1} \right)_{z_1=z, \,z_2=\overline{z}}, \qquad \displaystyle \frac{\partial g}{\partial \overline{z}}:=\left(\frac{\partial g}{\partial z_2} \right)_{z_1=z, \,z_2=\overline{z}}$.
\end{theorem}

\begin{definition}
A real valued function $f:\mathbb{C}\longrightarrow \mathbb{R}$ has a stationary point if
$$\frac{\partial}{\partial x}f(x,y)=\frac{\partial}{\partial y} f(x,y)=0, $$
where $z=x+iy.$ 
\end{definition}
It is easy to see that for such real valued function we have: 
$$\frac{\partial f}{\partial x}=\frac{\partial f}{\partial y} =0\Leftrightarrow \frac{\partial f}{\partial z}=0  \Leftrightarrow \frac{\partial f}{\partial \overline{z}}=0$$
which lead to the following result in~\cite{Brandwood}.

\begin{theorem}
Let $f:\mathbb{C}\longrightarrow \mathbb{R}$ be a real-valued function of a complex variable $z$ such that $f(z)=g(z,\overline{z})$ with $g:\mathbb{C}\times \mathbb{C}\longrightarrow \mathbb{R}$ is a function of two complex variables which is analytic with respect to each variable. Then, $f$ has a stationary point if and only if 
$$\frac{\partial g}{\partial \overline{z}}=0.$$
In a similar way, the condition below is also a necessary and sufficient condition for $f$ to have a stationary point 
$$\frac{\partial g}{\partial z}=0.$$ 
\end{theorem}

\subsection*{The Complex gradient operator}
\label{Cplx-grad}
  We now move to the case of several complex variables and use all the CR-calculus techniques described above. We define the complex gradient operators with respect to the $z$ and $\overline{z}$.

\begin{definition}
Let $\mathbf{z}=(z_1,...,z_n)\in\mathbb{C}^n$, with $z_k=a_k+ib_k$ for any $k=1,...,n$. Then, the complex gradient operator with respect to $\mathbf{z}$ is defined to be
\begin{equation}
\nabla_{\mathbf{z}}:=(\partial_{z_1},...,\partial_{z_n})^T,
\end{equation}
where $\partial_{z_l}$ are the complex derivatives with respect to the variable $z_l$ and $l=1,...,n$.

In a similar way, the gradient operator with respect to the conjugate is defined by \begin{equation}
\nabla_{\mathbf{\overline{z}}}:=(\partial_{\overline{z_1}},...,\partial_{\overline{z_n}})^T.
\end{equation}
\end{definition}
\begin{theorem}
Let $f:\mathbb{C}^n\longrightarrow \mathbb{R}$ be a real-valued function of a complex variable $z$ such that $f(z)=g(z,\overline{z})$ with $g:\mathbb{C}^n\times \mathbb{C}^n\longrightarrow \mathbb{R}$ is a function of two variables which is analytic with respect to each variable. Then, $f$ has a stationary point if and only if 
$$\nabla_\mathbf{z} g=0.$$
In a similar way, the condition below is also a necessary and sufficient condition for $f$ to have a stationary point 
$$\nabla_{\overline{\mathbf{z}}}g=0.$$

\end{theorem}

Here we summarize the properties of the gradient operators $\nabla_{\mathbf{z}}$ and $\nabla_\mathbf{\overline{z}}$, we leave the proof for the reader. 
\begin{proposition}
Let $a=(a_1,...,a_n)^T \in\mathbb{C}^n$ and $R$ be in $\mathbb{C}^{n\times n}$ respectively, then the following properties hold true:
\begin{enumerate}
\item $\nabla_{\mathbf{z}}(a^T\, \mathbf{z})=a$
\item $\nabla_{\mathbf{z}}(\overline{\mathbf{z}}^T a)=0$
\item $\nabla_{\mathbf{z}}(\overline{\mathbf{z}}^T R \mathbf{z})=(R)^T \overline{\mathbf{z}}$
\item $\nabla_{\mathbf{z}}(a^T \mathbf{z}+\mathbf{z}^Ta)=2a$
\item $\nabla_\mathbf{z}(a^T\overline{\mathbf{z}}+\overline{\mathbf{z}}^Ta)=0$
\item  $\nabla_\mathbf{z}(\overline{\mathbf{z}}^T\overline{\mathbf{z}})=0$
\item  $\nabla_\mathbf{z}(\overline{\mathbf{z}}^T\mathbf{z})=\overline{\mathbf{z}}$
\item  $\nabla_\mathbf{z}(\mathbf{z}^T\mathbf{z})=2\mathbf{z}$
\item $\nabla_{\overline{\mathbf{z}}}(\overline{a}^T \mathbf{z})=0$
\item $\nabla_{\overline{\mathbf{z}}}(\overline{\mathbf{z}}^T a)=a$
\item $\nabla_{\overline{\mathbf{z}}}(\overline{\mathbf{z}}^T R \mathbf{z})=R \mathbf{z}$.
\end{enumerate}
\end{proposition}

Now we review some applications of the complex CR-calculus.

\subsection*{Complex LMS algorithm}
\setcounter{equation}{0}
The authors of \cite{WMcB} extends to the complex case the well-known real Least Mean Square (LMS) algorithm which was originally introduced in the real case by Widrow and Hoff see \cite{WH1960}. In fact, at time $j$, let us consider the output given by $$y_j=x_j^Tw_j=w_j^Tx_j, $$
with $x_j=(x_{j,1},...,x_{j,n})^T$ and $w_j=(w_{j,1},...,w_{j,n})^T.$
Thus, we have 
$$y_j=\displaystyle \sum_{l=1}^nx_{j,l}w_{j,l}.$$ 
We denote by $d_j$ the desired response and the complex signal error is given by 
$$e_j=d_j-y_j=d_j-w_j^Tx_j.$$
Let us consider the complex LMS rule considered in Brandwood, see~\cite{Brandwood}:
\begin{equation}
w_{j+1}=w_j-\mu\nabla_{\overline{w_j}}(e_j\overline{e_j}),
\end{equation}
where $\mu>0$ and $$\nabla_{\overline{w_j}}:=\left(\partial_{\overline{w_{j,1}}},...,\partial_{\overline{w_{j,n}}}\right)^T.$$

\begin{theorem}
It holds that 
\begin{equation}\label{LMS}
w_{j+1}=w_j+\mu e_j\overline{x_j}.
\end{equation}
\end{theorem}
\begin{proof}
We have \begin{equation}
\nabla_{\overline{w_j}}(e_j\overline{e_j})=e_j\nabla_{\overline{w_j}}(\overline{e_j})+(\nabla_{\overline{w_j}} e_j)\overline{e_j},
\end{equation}
and the proof follows.
\end{proof}
\begin{remark}
The learning rule given by the formula \eqref{LMS} is the complex LMS algorithm proposed by Widrow and his collaborators.
\end{remark}

We have used the arguments of complex gradient operators and apply them to the study of bicomplex gradient ones. 
In particular, we write two least mean square algorithms (LMS) in the bicomplex settings.


\section*{Appendix B: Bicomplex CR - Calculus }

\begin{definition}
We  say that a function $f:\mathbb{BC}\longrightarrow \mathbb{BC}$ is \textbf{BC-R analytic (or BC-R regular)} if there exists an BC-analytic function of four bicomplex variables $$g:\mathbb{BC}\times \mathbb{BC}\times \mathbb{BC} \times \mathbb{BC}\longrightarrow \mathbb{BC}, (Z_1,Z_2,Z_3,Z_4)\mapsto g(Z_1,Z_2,Z_3,Z_4)$$ such that 
\begin{itemize}
\item[i)] $f(Z)=g(Z,\overline{Z},Z^*,Z^\dagger), \quad \forall Z\in \mathbb{BC}$,
\\
\item[ii)] $\displaystyle \frac{\partial f}{ \partial Z}=\left(\frac{\partial g}{\partial Z_1} \right)_{Z_1=Z, \,Z_2=\overline{Z},\,Z_3=Z^*\,Z_4=Z^\dagger}$ ,
\\
\item[iii)]  $\displaystyle \frac{\partial f}{ \partial \overline{Z}}=\left(\frac{\partial g}{\partial Z_2} \right)_{Z_1=Z, \,Z_2=\overline{Z},\,Z_3=Z^*\,Z_4=Z^\dagger}$,
\item[iv)] $\displaystyle \frac{\partial f}{ \partial Z^*}=\left(\frac{\partial g}{\partial Z_3} \right)_{Z_1=Z, \,Z_2=\overline{Z},\,Z_3=Z^*\,Z_4=Z^\dagger}$,
\item[v)] $\displaystyle \frac{\partial f}{ \partial Z^\dagger}=\left(\frac{\partial g}{\partial Z_4} \right)_{Z_1=Z, \,Z_2=\overline{Z},\,Z_3=Z^*\,Z_4=Z^\dagger}$.
\end{itemize}
\end{definition}
\begin{example}
The Finsler-type norm defined by $$f(Z)=|Z|_{\mathcal F}^4 := Z\bar{Z} Z^\ast Z^\dagger$$ is a BC-R analytic function on $\mathbb{BC}$ with $$ g(Z_1,Z_2,Z_3,Z_4)=Z_1Z_2Z_3Z_4.$$
We have $$\displaystyle \frac{\partial f}{ \partial Z}=\bar{Z} Z^\ast Z^\dagger, \quad  \frac{\partial f}{ \partial \overline{Z}}=Z Z^\ast Z^\dagger,$$
and $$\frac{\partial f}{ \partial Z^*}=Z\bar{Z}  Z^\dagger, \quad \frac{\partial f}{ \partial Z^\dagger}=Z\bar{Z} Z^\ast .$$
\end{example}
\begin{proposition}
Let $f,h:\mathbb{BC}\longrightarrow \mathbb{BC}$ be two BCR analytic functions and $\lambda \in\mathbb{BC} $. Then, the sum $f+h$ and multiplication $f\lambda$ are also BCR analytic.
\end{proposition}
\begin{proof}
Follows standard arguments.
\end{proof}
\begin{remark}
The set of all BCR analytic functions is a vector space over $\mathbb{BC}$ which is denoted by $\mathcal{H}_{CR}(\mathbb{BC})$.
\end{remark}

We can now prove Theorem~\ref{LeibBCn} which establishes the Bicomplex-gradient Leibniz rules, i.e.  for  $f$ and $g$ be bicomplex-valued $BC$-R analytic functions as in Equations ~\ref{G1}, ~\ref{G2}, and ~\ref{G3}
\begin{equation*}
\nabla_{\mathsf{Z}}(fg)=f\nabla_{\mathsf{Z}}(g)+ \nabla_{\mathsf{Z}}(f) g,
\end{equation*}

\begin{equation*}
\nabla_{\overline{\mathsf{Z}}}(fg)=f\nabla_{\overline{\mathsf{Z}}}(g)+ \nabla_{\overline{\mathsf{Z}}}(f) g,
\end{equation*}

\begin{equation*}
\nabla_{\mathsf{Z}^*}(fg)=f\nabla_{\mathsf{Z}^*}(g)+ \nabla_{\mathsf{Z}^*}(f) g.
\end{equation*}

\begin{proof}
We will prove the last formula, with respect to the bicomplex gradient-$*$ operator, using the idempotent decomposition, which lends itself to a more elegant proof.

We first write $\mathsf{Z}=\Lambda_1\mathbf{e}_1+\Lambda_2\mathbf{e}_2$, $f=f_1\mathbf{e}_1+f_2\mathbf{e}_2$ and  $g=g_1\mathbf{e}_1+g_2\mathbf{e}_2$.
Then, $$fg=f_1g_1\mathbf{e}_1+f_2g_2\mathbf{e}_2.$$
So, using the gradient operator representation in terms of the bicomplex variables $(\Lambda_1,\Lambda_2)$ we have 
\begin{align*}
\nabla_{\mathsf{Z}^*}(fg)&=(\nabla_{\overline{\Lambda_1}}\mathbf{e}_1+\nabla_{\overline{\Lambda_2}}\mathbf{e}_2)(fg)\\
&=\nabla_{\overline{\Lambda_1}}(fg)\mathbf{e}_1+\nabla_{\overline{\Lambda_2}}(fg)\mathbf{e}_2\\
\end{align*}
However, we note that 
$$\nabla_{\overline{\Lambda_1}}(fg)\mathbf{e}_1=\nabla_{\overline{\Lambda_1}}(f_1g_1)\mathbf{e}_1^2+\nabla_{\overline{\Lambda_1}}(f_2g_2)\mathbf{e}_2\cdot \mathbf{e}_1.$$
Thus, using the fact that $\mathbf{e}_1^2=\mathbf{e}_1$ and $\mathbf{e}_2\cdot \mathbf{e}_1=0$ we obtain
$$\nabla_{\overline{\Lambda_1}}(fg)\mathbf{e}_1=\nabla_{\overline{\Lambda_1}}(f_1g_1)\mathbf{e}_1,$$
and similarly:
$$\nabla_{\overline{\Lambda_2}}(fg)\mathbf{e}_2=\nabla_{\overline{\Lambda_2}}(f_2g_2)\mathbf{e}_2.$$
Since the Leibniz rule holds in the classical complex case this finish to the proof of the rule for this gradient operator since:
\begin{equation*}
\nabla_{\mathsf{Z}^*}(fg)=\nabla_{\overline{\Lambda_1}}(f_1g_1)\mathbf{e}_1+\nabla_{\overline{\Lambda_2}}(f_2g_2)\mathbf{e}_2.
\end{equation*}

Using the properties of the idempotent elements $\mathbf{e}_1$ and $\mathbf{e}_2$ the proof follows and we have:
\begin{align*}
\nabla_{\mathsf{Z}^*}(fg)=f\nabla_{\mathsf{Z}^*}(g)+ \nabla_{\mathsf{Z}^*}(f) g.
\end{align*}
Similar arguments can be used for the other gradient operators.
\end{proof}

We formulate a hypothesis proposal for the bicomplex stationary point conditions:
\begin{theorem}
Let $f:\mathbb{BC}^n\longrightarrow \mathbb{R}$ be a real-valued function of a bicomplex variable $\mathsf{Z}$ such that $f(\mathsf{Z})=g(\mathsf{Z},\overline{\mathsf{Z}},\mathsf{Z}^*, \mathsf{Z}^{\dagger})$ with $g:(\mathbb{BC}^n)^4\longrightarrow \mathbb{R}$ is a function of four variables which is analytic with respect to each variable. Then, $f$ has a stationary point if and only if one of the following equivalent conditions is satisfied:
\begin{enumerate}
\item[i)] $\nabla_\mathsf{Z} g=0,$
\item[ii)] $\nabla_{\overline{\mathsf{Z}}}g=0,$
\item[iii)] $\nabla_{\mathsf{Z}^*}g=0,$
\item[iv)]$\nabla_{\mathsf{Z}^{\dagger}}g=0.$
\end{enumerate}
\end{theorem}
\begin{proof}
We write $f(\mathsf{Z})=f_1(\mathsf{Z})+\mathbf{i}f_2(\mathsf{Z})+\mathbf{j}f_3(\mathsf{Z})+\mathbf{k}f_4(\mathsf{Z})$. Then, since $f=g$ is real valued, we have $f=f_1$ and it is clear that the conditions above are all equivalent in this case. Moreover, $f$ has a stationary point if and only one of these conditions hold. In fact, since $f$ is real valued and using the bicomplex gradient operators definition we have $\nabla_\mathsf{Z} f=\nabla_\mathsf{Z}f_1 =0 $ if and only if $\nabla_{\mathbf{z_1}}f_1=\nabla_{\mathbf{z_2}}f_1=0$ and then by the Brandwood complex result this is equivalent to   $\nabla_{\overline{\mathbf{z_1}}}f_1=\nabla_{\overline{\mathbf{z_2}}}f_1=0$. Hence, in order to have a stationary point it is enough to have $\nabla_\mathsf{Z} g=0$ or $\nabla_{\overline{\mathsf{Z}}}g=0$ or $\nabla_{\mathsf{Z}^*}g=0$ or $\nabla_{\mathsf{Z}^{\dagger}}g=0.$
\end{proof}


\section*{Appendix C: Proofs of the two BLMS Algorithms}
\label{Proofs}
\setcounter{equation}{0}


In this appendix we add the proofs of the two BLMS algorithms. For more details we point the reader to the following references~\cite{ADV-BCR,ADV-BCR2}.


\subsection*{Proofs of Theorems ~\ref{LMSR1} and ~\ref{LMS1decomp}}
We now include a proof of the first BLMS algorithm, i.e. Theorem~\ref{LMSR1} which states that 
the learning rule at time $\ell$ of the first BLMS algorithm has the following explicit expression:
 \begin{equation*}
\mathsf{W}_{\ell+1}=\mathsf{W_\ell}+2\mu \mathsf{E}_\ell\mathsf{X}_\ell^*.
\end{equation*}

\begin{proof}
We write the first bicomplex LMS algorithm at time $\ell$ as follows $$\mathsf{W}_{\ell+1}=\mathsf{W_\ell}-\mu \nabla_{\mathsf{W}_\ell^*}(\mathsf{E}_\ell \mathsf{E}_\ell^*).$$
We have also
$$
\mathsf{E}_\ell \mathsf{E}_\ell^*=\mathsf{D}_\ell \mathsf{D}_\ell^*-\mathsf{D}_\ell \mathsf{Y}_\ell^*-\mathsf{Y}_\ell \mathsf{D}_\ell^*+\mathsf{Y}_\ell \mathsf{Y}_\ell^*.
$$
Then, starting form \eqref{bicompW1} it is clear that we only have to compute the quantity $\nabla_{\mathsf{W}_\ell^*}(\mathsf{E}_\ell \mathsf{E}_\ell^*)$. We apply Theorem \ref{LeibBCn} and get 
$$\nabla_{\mathsf{W}_\ell^*}(\mathsf{E}_\ell\mathsf{E}_\ell^*)=\mathsf{E}_\ell \nabla_{\mathsf{W}_\ell^*}(\mathsf{E}_\ell^*)+\mathsf{E}_\ell^* \nabla_{\mathsf{W}_\ell^*}(\mathsf{E}_\ell).$$
First, we note that the error $\mathsf{E}_\ell$ is given by
$$\displaystyle \mathsf{E}_\ell:=\mathsf{D}_\ell-\mathsf{Y}_\ell=\mathsf{D}_\ell- \sum_{k=1}^n\mathsf{X}_{\ell, k} \mathsf{W}_{\ell, k}$$
So, using the properties of the operator $\nabla_{\mathsf{Z}^*}$ we have 
\begin{align*}
\nabla_{\mathsf{W}_\ell^*}(\mathsf{E}_\ell)&=-\sum_{k=1}^n\mathsf{X}_{\ell, k} \nabla_{\mathsf{W}_\ell^*}(\mathsf{W}_{\ell, k})\\
&=0.\\
\end{align*}
On the other hand, since 
$$\displaystyle \mathsf{E}_\ell^*=\mathsf{D}_\ell^*-\mathsf{Y}_\ell^*=\mathsf{D}_\ell^*- \sum_{k=1}^n\mathsf{X}_{\ell, k}^* \mathsf{W}_{\ell, k}^*,$$

 we have the following

\begin{align*}
\nabla_{\mathsf{W}_\ell^*}(\mathsf{E}_\ell^*)&=-\sum_{k=1}^n\mathsf{X}_{\ell, k}^* \nabla_{\mathsf{W}_\ell^*}(\mathsf{W}_{\ell, k}^*).
\end{align*}
We observe the following fact

$$\partial_{\mathsf{W}_{\ell,k}^*}(\mathsf{W}_{\ell,k}^*)=2, \quad \partial_{\mathsf{W}_{\ell,s}^*}(\mathsf{W}_{\ell,k}^*)=0, \qquad \forall k,s=1,...,n, s\neq k.$$
In particular, we note that for every $k=1,..., n$ we have 

\begin{equation*}
\mathsf{X}_{\ell, k}^* \nabla_{\mathsf{W}_\ell^*}(\mathsf{W}_{\ell, k}^*) = 
\begin{pmatrix}
\mathsf{X}_{\ell, k}^* \partial_{\mathsf{W}_{\ell,1}^*}(\mathsf{W}_{\ell,k}^*)&\\
\vdots&\\
\mathsf{X}_{\ell, k}^* \partial_{\mathsf{W}_{\ell,k}^*}(\mathsf{W}_{\ell,k}^*)&\\
\mathsf{X}_{\ell, k}^* \partial_{\mathsf{W}_{\ell,k+1}^*}(\mathsf{W}_{\ell,k}^*) & \\
\vdots &\\
\mathsf{X}_{\ell, k}^* \partial_{\mathsf{W}_{\ell,n}^*}(\mathsf{W}_{\ell,k}^*)&\\
& 
\end{pmatrix}
= \begin{pmatrix}
0 & \\
\vdots &  \\

2 \mathsf{X}_{\ell, k}^*& \\
0&\\
\vdots  & \\
0 & \\
& 
\end{pmatrix}.
\end{equation*}

So, we obtain \begin{equation}
\sum_{k=1}^n\mathsf{X}_{\ell, k}^* \nabla_{\mathsf{W}_\ell^*}(\mathsf{W}_{\ell, k}^*)=2 \begin{pmatrix}
\mathsf{X}_{\ell,1} ^*& \\
\vdots &  \\

\mathsf{X}_{\ell, k}^*& \\
\vdots  & \\
\mathsf{X_{\ell,n}^*} & \\
& 
\end{pmatrix}= 2\mathsf{X}_\ell^*.
\end{equation}
Thus, it follows that at time $\ell$ we have
\begin{align*}
\nabla_{\mathsf{W}_\ell^*}(\mathsf{E}_\ell^*)&=-\sum_{k=1}^n\mathsf{X}_{\ell, k}^* \nabla_{\mathsf{W}_\ell^*}(\mathsf{W}_{\ell, k}^*)\\
&=-2\mathsf{X}_\ell^*.\\
\end{align*}
Hence, to sum up we have
$$\nabla_{\mathsf{W}_\ell^*}(\mathsf{E}_\ell)=0,$$
and
$$\nabla_{\mathsf{W}_\ell^*}(\mathsf{E}_\ell^*)=-2 \mathsf{X}_\ell^*.$$
We obtain the final step of our proof:
$$\nabla_{\mathsf{W}_\ell^*}(\mathsf{E}_\ell\mathsf{E}_\ell^*)=\mathsf{E}_\ell \nabla_{\mathsf{W}_\ell^*}(\mathsf{E}_\ell^*)=-2\mathsf{E}_\ell \mathsf{X}_\ell^*,$$
which yields at time $\ell$ the first bicomplex LMS learning rule given by 
$$\mathsf{W}_{\ell+1}=\mathsf{W_\ell}+2\mu \mathsf{E}_\ell \mathsf{X}_\ell^*.$$
\end{proof}

The proof of Theorem~\ref{LMS1decomp} follows, using the decomposition of the bicomplex weights, error and input which are given respectively by $$\mathsf{W}_\ell=w_{\ell,1}\mathbf{e}_1+w_{\ell,2}\mathbf{e}_2, \mathsf{E}_\ell=e_{\ell,1}\mathbf{e}_1+e_{\ell,2}\mathbf{e}_2,\quad \text{ and }\mathsf{X}_\ell=x_{\ell,1}\mathbf{e}_1+x_{\ell,2}\mathbf{e}_2.$$ We prove that, at time $\ell$, the learning rule of the first bicomplex LMS algorithm can be expressed in terms of two complex LMS algorithms. 

\begin{align*}
\mathsf{W}_{\ell+1}&=(w_{\ell,1}+2\mu e_{\ell,1}\overline{x_{\ell,1}})\mathbf{e}_1+(w_{\ell,2}+2\mu e_{\ell,2}\overline{x_{\ell,2}})\mathbf{e}_2\\
&=w_{\ell+1,1}\mathbf{e}_1+w_{\ell+1,2}\mathbf{e}_2.
\end{align*}

\begin{proof}
We use the expression of the first bicomplex LMS algorithm obtained in Theorem \ref{LMSR1} and get 

\begin{align*}
\mathsf{W}_{\ell+1}&=\mathsf{W_\ell}+2\mu \mathsf{E}_\ell\mathsf{X}_\ell^*\\
&=(w_{\ell,1}\mathbf{e}_1+w_{\ell,2}\mathbf{e}_2)+2\mu (e_{\ell,1}\mathbf{e}_1+e_{\ell,2}\mathbf{e}_2)(x_{\ell,1}\mathbf{e}_1+x_{\ell,2}\mathbf{e}_2)^*\\
&=(w_{\ell,1}\mathbf{e}_1+w_{\ell,2}\mathbf{e}_2)+2\mu (e_{\ell,1}\mathbf{e}_1+e_{\ell,2}\mathbf{e}_2)(\overline{x_{\ell,1}}\mathbf{e}_1+\overline{x_{\ell,2}}\mathbf{e}_2)\\
\end{align*}
So, applying the product formula and properties of the $*$-bicomplex conjugate we have

\begin{align*}
\mathsf{W}_{\ell+1}&=(w_{\ell,1}\mathbf{e}_1+w_{\ell,2}\mathbf{e}_2)+2\mu (e_{\ell,1}\overline{x_{\ell,1}}\mathbf{e}_1+e_{\ell,2}\overline{x_{\ell,2}}\mathbf{e}_2)\\
&=(w_{\ell,1}+2\mu e_{\ell,1}\overline{x_{\ell,1}})\mathbf{e}_1+(w_{\ell,2}+2\mu e_{\ell,2}\overline{x_{\ell,2}})\mathbf{e}_2\\
&=w_{\ell+1,1}\mathbf{e}_1+w_{\ell+1,2}\mathbf{e}_2.\\
\end{align*}
Finally, the first BLMS algorithm can be represented by the following two complex LMS algortihms given by 

$$w_{\ell+1,1}=w_{\ell,1}+2\mu e_{\ell,1}\overline{x_{\ell,1}},$$
and $$w_{\ell+1,2}=w_{\ell,1}+2\mu e_{\ell,2}\overline{x_{\ell,2}}.$$
\end{proof}

\subsection*{Proofs of Theorems~\ref{SLMSalgo} and~\ref{LMS2decomp}}

We now include the proof of the second BLMS algorithm we found, i.e. Theorem~\ref{SLMSalgo}, which states that
the learning rule at time $\ell$ of the second bicomplex LMS algorithm has the following explicit expression: \begin{equation*}
\mathsf{W}_{\ell+1}=\mathsf{W_\ell}+2\mu \mathsf{E}_\ell\overline{\mathsf{X}_\ell}.
\end{equation*}

\begin{proof}
We have
\begin{equation}
\mathsf{E}_\ell \overline{\mathsf{E}_\ell}=\mathsf{D}_\ell \overline{\mathsf{D}_\ell}-\mathsf{D}_\ell \overline{\mathsf{Y}_\ell}-\mathsf{Y}_\ell \overline{\mathsf{D}_\ell}+\mathsf{Y}_\ell \overline{\mathsf{Y}_\ell}.
\end{equation}
Starting form \eqref{bicompW} it is clear that we only have to compute the quantity $\nabla_{\overline{\mathsf{W}_\ell}}(\mathsf{E}_\ell \overline{\mathsf{E}_\ell})$. To this end, we apply Theorem \ref{LeibBCn} and get 
$$\nabla_{\overline{\mathsf{W}_\ell}}(\mathsf{E}_\ell\overline{\mathsf{E}_\ell})=\mathsf{E}_\ell \nabla_{\overline{\mathsf{W}_\ell}}(\overline{\mathsf{E}_\ell})+\overline{\mathsf{E}_\ell} \nabla_{\mathsf{W}_\ell^*}(\mathsf{E}_\ell).$$
First, we note that the error $\mathsf{E}_\ell$ is given by
$$\displaystyle \mathsf{E}_\ell:=\mathsf{D}_\ell-\mathsf{Y}_\ell=\mathsf{D}_\ell- \sum_{k=1}^n\mathsf{X}_{\ell, k} \mathsf{W}_{\ell, k}$$
So, using the properties of the operator $\nabla_{\overline{\mathsf{Z}}}$ we have 
\begin{align*}
\nabla_{\overline{\mathsf{W}_\ell}}(\mathsf{E}_\ell)&=-\sum_{k=1}^n\mathsf{X}_{\ell, k} \nabla_{\overline{\mathsf{W}_\ell}}(\mathsf{W}_{\ell, k})\\
&=0.\\
\end{align*}
On the other hand, since 
$$\displaystyle \overline{\mathsf{E}_\ell}=\overline{\mathsf{D}_\ell}-\overline{\mathsf{Y}_\ell}=\overline{\mathsf{D}_\ell}- \sum_{k=1}^n\overline{\mathsf{X}_{\ell, k}} \overline{\mathsf{W}_{\ell, k}},$$

and 
$$\partial_{\overline{\mathsf{W}_{\ell,k}}}(\overline{\mathsf{W}_{\ell,k}})=2, \quad \partial_{\overline{\mathsf{W}_{\ell,s}}}(\overline{\mathsf{W}_{\ell,k}})=0, \qquad \forall k,s=1,...,n, s\neq k.$$

Following a similar reasoning as for the first BLMS algorithm we can develop computations with respect to the  bicomplex $bar$-gradient operator and get

\begin{align*}
\nabla_{\overline{\mathsf{W}_\ell}}(\overline{\mathsf{E}_\ell})&=-\sum_{k=1}^n\overline{\mathsf{X}_{\ell, k}} \nabla_{\overline{\mathsf{W}_\ell}}(\overline{\mathsf{W}_{\ell, k}})\\
&=-2\overline{\mathsf{X}_\ell}.\\
\end{align*}
Hence, to sum up we have
$$\nabla_{\overline{\mathsf{W}_\ell}}(\mathsf{E}_\ell)=0,$$
and
$$\nabla_{\overline{\mathsf{W}_\ell}}(\overline{\mathsf{E}_\ell})=- 2\overline{\mathsf{X}_\ell}.$$
Finally, we obtain 
$$\nabla_{\overline{\mathsf{W}_\ell}}(\mathsf{E}_\ell\overline{\mathsf{E}_\ell})=\mathsf{E}_\ell \nabla_{\overline{\mathsf{W}_\ell}}\overline{\mathsf{E}_\ell}=-2\mathsf{E}_\ell \overline{\mathsf{X}_\ell},$$
which yields to the second BLMS learning rule given by $$\mathsf{W}_{\ell+1}=\mathsf{W_\ell}+2\mu \mathsf{E}_\ell\overline{\mathsf{X}_\ell}.$$
\end{proof}

Similarly, we now prove Theorem~\ref{LMS2decomp}, where we consider the decomposition of the bicopmplex weights, error and input which are given respectively by $$\mathsf{W}_\ell=w_{\ell,1}\mathbf{e}_1+w_{\ell,2}\mathbf{e}_2, \quad \mathsf{E}_\ell=e_{\ell,1}\mathbf{e}_1+e_{\ell,2}\mathbf{e}_2,\quad \text{ and }\quad \mathsf{X}_\ell=x_{\ell,1}\mathbf{e}_1+x_{\ell,2}\mathbf{e}_2.$$ Then, at time $\ell$ the learning rule of the second BLMS algorithm can be expressed in terms of two complex LMS algorithms as follows: 

\begin{align*}
\mathsf{W}_{\ell+1}&=(w_{\ell,1}+2\mu e_{\ell,1}\overline{x_{\ell,2}})\mathbf{e}_1+(w_{\ell,2}+2\mu e_{\ell,2}\overline{x_{\ell,1}})\mathbf{e}_2\\
&=w_{\ell+1,1}\mathbf{e}_1+w_{\ell+1,2}\mathbf{e}_2.
\end{align*}

\begin{proof}
We use the expression of the second bicomplex LMS algorithm obtained in Theorem \ref{SLMSalgo} and get 

\begin{align*}
\mathsf{W}_{\ell+1}&=\mathsf{W_\ell}+2\mu \mathsf{E}_\ell\overline{\mathsf{X}_\ell}\\
&=(w_{\ell,1}\mathbf{e}_1+w_{\ell,2}\mathbf{e}_2)+2\mu(e_{\ell,1}\mathbf{e}_1+e_{\ell,2}\mathbf{e}_2)\overline{(x_{\ell,1}\mathbf{e}_1+x_{\ell,2}\mathbf{e}_2)}\\
&=(w_{\ell,1}\mathbf{e}_1+w_{\ell,2}\mathbf{e}_2)+2\mu (e_{\ell,1}\mathbf{e}_1+e_{\ell,2}\mathbf{e}_2)(\overline{x_{\ell,1}}\mathbf{e}_2+\overline{x_{\ell,2}}\mathbf{e}_1).\\
\end{align*}
So, applying the product rule and properties of the $bar$-bicomplex conjugate, such as $\overline{\mathbf{e}_1}=\mathbf{e}_2$ and $\overline{\mathbf{e}_2}=\mathbf{e}_1$, we obtain

\begin{align*}
\mathsf{W}_{\ell+1}&=(w_{\ell,1}\mathbf{e}_1+w_{\ell,2}\mathbf{e}_2)+2\mu (e_{\ell,1}\overline{x_{\ell,2}}\mathbf{e}_1+e_{\ell,2}\overline{x_{\ell,1}}\mathbf{e}_2)\\
&=(w_{\ell,1}+2\mu e_{\ell,1}\overline{x_{\ell,2}})\mathbf{e}_1+(w_{\ell,2}+2\mu e_{\ell,2}\overline{x_{\ell,1}})\mathbf{e}_2\\
&=w_{\ell+1,1}\mathbf{e}_1+w_{\ell+1,2}\mathbf{e}_2.\\
\end{align*}
Finally, the second BLMS algorithm can be represented by the following two complex LMS algortihms given by 

$$w_{\ell+1,1}=w_{\ell,1}+2\mu e_{\ell,1}\overline{x_{\ell,2}},$$
and $$w_{\ell+1,2}=w_{\ell,2}+2\mu e_{\ell,2}\overline{x_{\ell,1}}.$$
\end{proof}

{\it Data sharing not applicable to this article as no datasets were generated or analyzed during the current study. We also certify that this work does not have any conflicts of interest.}

\bibliographystyle{plain}

\end{document}